# A multi-branch convolutional neural network for detecting double JPEG compression

**Bin Li, Hu Luo, Haoxin Zhang, Shunquan Tan, Zhongzhou Ji**
*Shenzhen Key Lab of Media Information Security, Shenzhen University, P. R. China*

**ABSTRACT**

Detecting double JPEG compression is important to forensics analysis. A few methods were proposed based on convolutional neural networks (CNNs). These methods only accept inputs from pre-processed data, such as histogram features and/or decompressed images. In this paper, we present a CNN solution by using raw DCT (discrete cosine transformation) coefficients from JPEG images as input. Considering the DCT sub-band nature in JPEG, a multiple-branch CNN structure has been designed to reveal whether a JPEG format image has been doubly compressed. Comparing with previous methods, the proposed method provides end-to-end detection capability. Extensive experiments have been carried out to demonstrate the effectiveness of the proposed network.

## INTRODUCTION

With the tremendous evolution of Internet and online social media, multimedia data (especially images) have been pervasively used due to their abilities to show contents intuitively, exchange information conveniently, and entertain the public ubiquitously. Meanwhile, common people can edit images or videos with handy software; while human eyes can hardly notice the changes. Consequently, increasing concerns are drawn on the security issues regarding the originality and the authenticity of multimedia data, which is the main task of digital forensics.

JPEG (Wallace, 1992) images are the most common ones generated from cameras and widely spread on the Internet. Therefore, there is a high probability that a forger would choose JPEG images to edit, such as performing copy-move or splicing operations, and afterwards performing JPEG recompression for distribution. Detection of double JPEG compression may provide useful information to reveal the traces of image edition or fabricated manipulation.

Various kinds of approaches have been proposed to perform double JPEG detection based on extracting handcrafted features (Lukas *et al*., 2003; Popescu & Farid, 2004; Fu *et al*., 2007; Li *et al*., 2008; Amerini *et al.* 2014; Pasquini *et al.*, 2014; Taimori *et al.*, 2016). It is generally assumed that the AC (alternating current) DCT (discrete cosine transform) coefficients of a JPEG image follow Laplacian distribution, and their first significant digits conform Benford's law. When the image has been JPEG compressed twice with a different quantization matrix, the distribution of DCT coefficients is deviated from the statistical model. As a result, handcrafted features are often extracted based on the statistics of DCT coefficients. (Lukas *et al*., 2003) discovered that the histogram of DCT coefficients in a doubly compressed JPEG image fluctuate periodically. The phenomenon, called double quantization (DQ) effect, has been mathematically verified by (Popescu &Farid, 2004). Based on the observation in (Fu *et al*., 2007) that the DCT coefficients of singly JPEG compressed images follows the Benford's law, while that of doubly compressed images violating it, (Li *et al.*, 2008) designed mode-based first digit features (MBFDF) to effectively detect double JPEG compression. They extracted the first significant digits (0~9) of DCT coefficients in the first 20 AC sub-bands, and concatenated them as a 180-D feature vector. (Amerini *et al.* 2014) took the first digits of 2, 5, and 7 in the first 9 AC sub-bands to reduce the feature dimension to 27-D. (Pasquini *et al.*, 2014) proposed a statistical model based on Benford-Fourier coefficients. (Taimori *et al.*, 2016) developed a quantization table unaware method for double JPEG compression detection based on Benford's law by using all non-zero and zero AC coefficients.

It is tempting that data-driven approaches, such as those based on deep learned features (LeCun *et al*., 2015; Simonyan & Zisserman, 2014; Szegedy *et al*., 2015; He *et al*., 2016), may explore hierarchy data representation, and automatically expose statistical deviation without knowing much of the details of image manipulation. (Wang & Zhang, 2016; Amerini *et al.*, 2017) have proposed effective CNNs for double JPEG compression. However, as shown in Fig. 1 with the overview of network architecture and detailedly described in the next section, the input of their CNNs is preprocessed by extracting the histogram of DCT coefficients, instead of directly using the raw DCT coefficients. As we know, histogram is a kind of first-order statistics. Such a preprocessing stage has already provided some handcrafted information for double JPEG compression, as implicitly shown in (Lukas *et al*., 2003). As a result, these CNNs may lack of end-to-end feature extraction capability.

To the best of our knowledge, there is rare work taking the raw JPEG DCT coefficients as the CNN input. In this paper, we propose a novel CNN for double JPEG detection. Different from the two frameworks presented in (Wang & Zhang, 2016; Amerini *et al.*, 2017), the proposed CNN-based architecture does not use the histogram features as input and provides end-to-end detection capability. The proposed network architecture, called JPEG-CNN, has the following highlights.
- The network takes raw DCT coefficients of JPEG image as input, without any handcrafted preprocessing operation such as histogram extraction.
- The network adapts to JPEG DCT sub-band structure by using a multi-branch architecture so that the intricate relationship of DCT coefficients within an AC sub-band and across AC sub-bands can be taken into consideration.
- Some additional layers, such as ABS (absolute value) layers and BN (batch normalization) layers, are utilized to regularize the network so as to improve its performance.

The rest of the paper is organized as follows. Some related works based on CNNs are reviewed in the next section. Our proposed CNN-based architecture for double JPEG compression is presented in the third section, where the design philosophy is introduced and each component is described in detail. Experimental results and performance analysis are given in the fourth section to demonstrate the effectiveness of the proposed JPEG-CNN. Concluding remarks are drawn in the final section.

# RELATED WORK

Plenty of works have been proposed to address double JPEG compression detection by using handcrafted features (Malviya & Naskar, 2014; Nowroozi & Zakerolhosseini, 2015; Shang *et al.*, 2016; Taimori *et al.*, 2017). But there are only a few attempts on applying CNNs (convolutional neural networks) for double JPEG compression detection.

(Wang & Zhang, 2016) extracted the histogram of DCT coefficients in the range of [-5, 5] from the first 9 AC sub-bands in zigzag order to form a 99-D feature vector as CNN input. Their CNN consists of two convolutional blocks and three fully connected layers. In each convolutional block, one dimensional convolution and one dimensional max pooling is used to obtain feature maps. ReLU (rectified linear unit) activation functions are used on each feature map and dropout operations are used in fully connected layers for regularization. The final fully connected layer uses a softmax connection to calculate the distribution of each class (singly compressed or doubly compressed). Such a one dimensional CNN provides better performance than the prior works (Bianchi & Piva, 2012; Amerini *et al.*, 2014).

(Amerini *et al.*, 2017) took a step forward. First, for the one-dimensional ***frequency-domain CNN*** as in (Wang & Zhang, 2016), they expanded the input to 909-D by using the histogram in the range of [-50, 50]. Second, they used a two-dimensional ***spatial-domain CNN*** with four convolutional blocks and two fully connected layers, where a convolutional block involves convolution and ReLU activation. Max pooling and dropout operations are used in the second and the fourth convolutional blocks. The final fully connected layer uses a softmax connection. Image pixel values (normalized in 0 to 1) from three color channels are served as the input of the spatial-domain network. Finally, by concatenating two kinds of feature maps of the first fully connected layer respectively from the frequency-domain CNN and the spatial-domain CNN, a ***multi-domain CNN*** is finally constructed. Experiments show that the spatial-domain CNN performs less effective than the frequency-domain CNN, and the multi-domain CNN performs better than both individual networks.

*Figure 1. An overview of the three CNN-based architectures.*

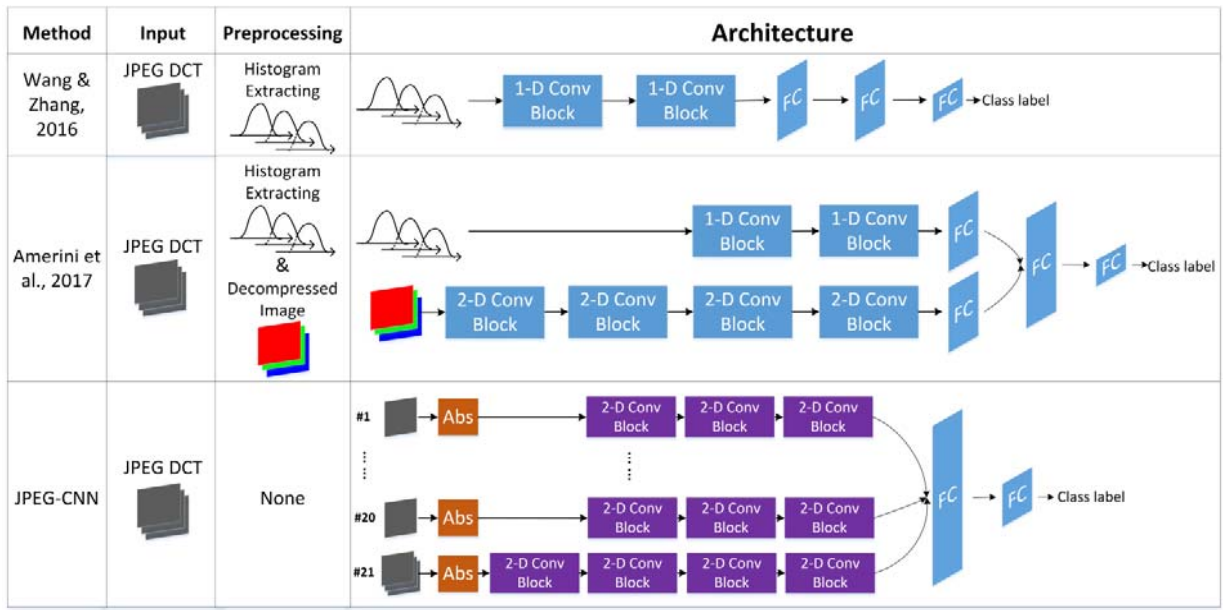

*Figure 2. The detailed network architecture of the proposed JPEG-CNN.*

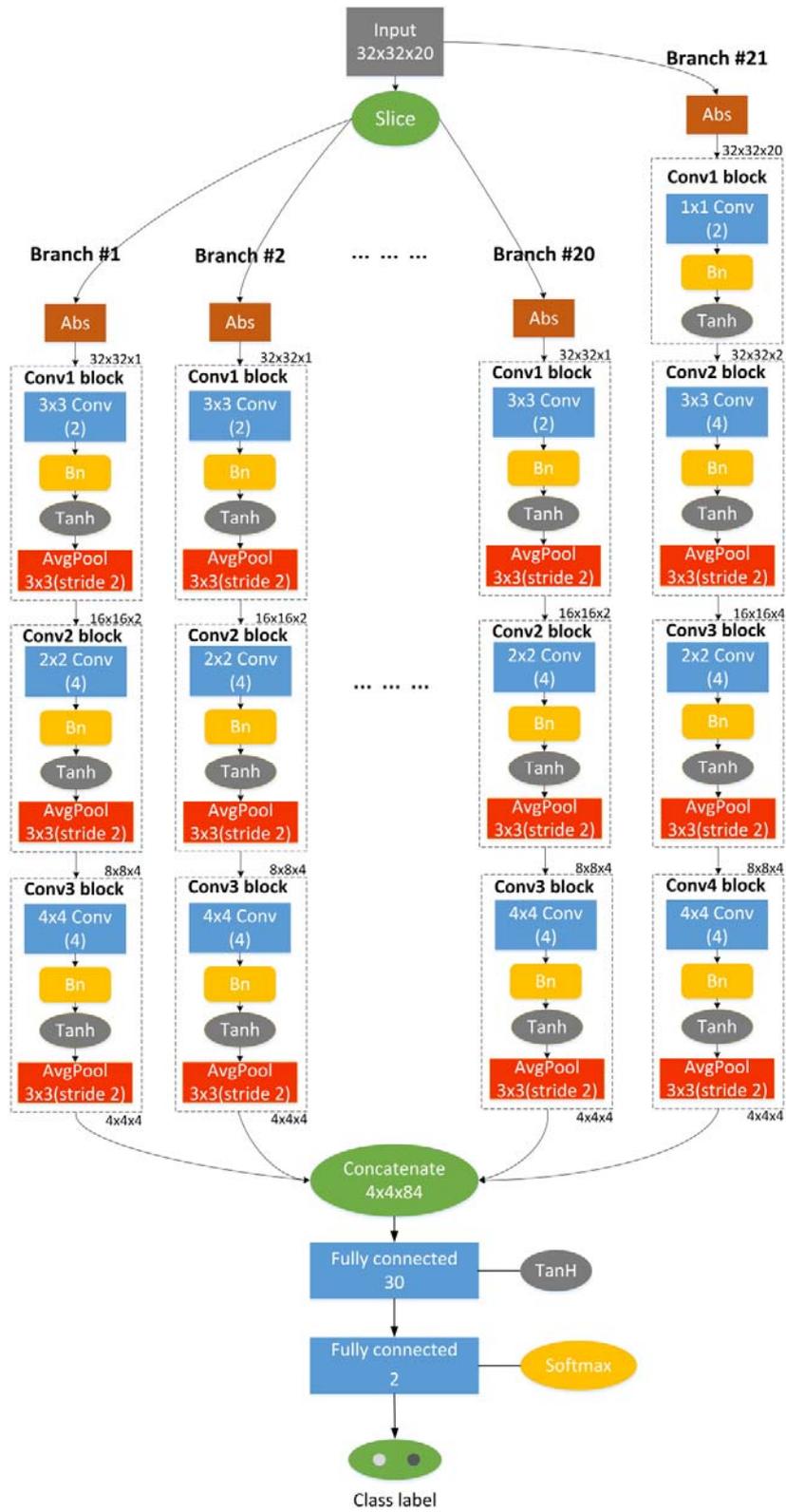

# JPEG-CNN ARCHITECTURE

The prior works (Wang & Zhang, 2016; Amerini *et al.*, 2017) by using CNN for detecting double JPEG compression either use histogram features or/and decompressed spatial representation as input. In this paper, we propose a novel CNN for double JPEG detection by taking the raw DCT coefficients of a JPEG image as input. It adapts to JPEG DCT sub-bands by exploiting a multi-branch structure. In the following subsections, we will first introduce the input of the proposed JPEG-CNN. Then we give the overview of the network architecture. Later, we will explain the details of each CNN layers. Finally, we briefly describe the loss function and learning procedure.

**Network Input**

A JPEG image has one DC (direct current) sub-band and 63 AC sub-bands by using 8×8 block DCT transform. During compression, each sub-band is associated with a quantization step defined by JPEG quality factor. If the quantization steps are different in two successive JPEG compressions, DQ effect (Lukas *et al.*, 2003) may occur in the corresponding sub-band. Generally low-frequency AC sub-bands have more prominent DQ effects than DC sub-band and high-frequency AC sub-bands. Consequently, previous methods usually extract handcrafted features from low-frequency AC sub-bands. In the proposed JPEG-CNN, we use the DCT coefficients of the first 20 AC sub-bands in zigzag order as input. Take an image of size 256×256 as example. There are 32×32 (256/8 ×256/8) DCT coefficients in each sub-band. By using 20 sub-bands, the image is represented by a 32×32×20 (height×width×depth) tensor, serving as the input of the proposed JPEG-CNN.

**Network Architecture Overview**

In order to adapt the CNN to the special multiple sub-bands data structure of JPEG DCT coefficients, we design a network structure with multiple branches, as shown in Fig. 2. With 20 sub-bands as input, we build 21 branches. Specifically, Branch #1 to Branch #20 respectively take the DCT coefficients from the corresponding AC sub-band from the 1st to the 20th as input, where each input is represented by a height×width×1 tensor. A special branch, namely, Branch #21, takes all AC sub-bands from 1st to the 20th as its input. We regard Branch #1 to Branch #20 as *intra sub-band branches*, and Branch #21 as a *inter sub-band branch*. In this way, the intricate relationship of DCT coefficient within each AC sub-band and across all the AC sub-bands can be taken into consideration.

Branch #1 to Branch #20 use identical network structure, which consists of an ABS layer and three convolutional blocks. In each convolutional block, a stack of convolution, batch normalization (Ioffe & Szegedy, 2015), TanH (Tan Hyperbolic) activation (Kalman & Kwasny, 1992), and average pooling is involved. The multi-branch structure for intra sub-bands looks like the ResNext structure (Xie *et al.*, 2016), in which a multi-path network is used. However, each path in ResNext shares the same input, while each intra sub-band branch in JPEG-CNN does not.

Branch #21 consists of an ABS layer and four groups of convolutional blocks. The first convolutional block uses 1×1 convolution to fuse the information from different sub-bands. It does not use pooling operation so that its output feature maps have the same height and width as the input of the intra sub-band branch. The last three convolutional blocks are identical to those in the intra sub-band branch, just with a slightly different number of convolutional kernels due to the different depths of the input and output feature maps. The convolutional kernel sizes and the output feature maps are elaborated in Table 1.

## Details of Network Layers
- **Slice layer**

Denote $D^{h \times w \times d}$ as a three dimensional tensor of size $h \times w \times d$, where $h$, $w$ and $d$, are the height, width, and depth, respectively. Denote $D^{h \times w \times d}(i)$ as the $i$-th slice of the tensor, which is in fact a two dimensional matrix of size $h \times w$. Assume a JPEG image is of size $h \times w$, the input of the proposed JPEG-CNN is denoted by $I^{x \times y \times z}$, where $x = h/8$, $y = w/8$ and $z = 20$. We use a slice layer to obtain the input of $k$-th intra sub-band branch by taking out the $k$-th slice of $I^{x \times y \times z}$, while we use $I^{x \times y \times z}$ directly as the input of the inter sub-band branch. In other words, the input of the $k$-th branch $I^{(k)}$ is:

$$I^{(k)} = \begin{cases} I^{x \times y \times z}(k), & k = 1, 2, ..., 20, \\ I^{x \times y \times z}, & k = 21. \end{cases} \quad (1)$$

- **ABS layer**

Before feeding the input to convolutional blocks, an ABS layer is employed to take the amplitudes of the DCT coefficients, as shown in equation (2).

$$\hat{I}^{(k)} = \left| I^{(k)} \right|, \quad k = 1, 2, ..., 21. \quad (2)$$

Such a process helps the network to get rid of the irrelevant sign information, which may not be useful in detecting double JPEG compression. Experiments verify the claim, as shown in the next section.

- **Convolutional layer**

A convolutional layer uses a set of two-dimensional kernels to convolve with input feature maps. The parameters of the convolution kernels $w$, or called weights, are updated by the BP (backward propaganda) algorithm as described in the next subsection.

Denote $R_i$ the $i$-th input feature map with size $M_1 \times M_2$, and $F_j$ the $j$-th output feature map. The convolution is performed as

$$F_j^{(k)}(u,v) = \sum_{i=1}^{D_1} \left( \sum_{m=0}^{K_1-1} \sum_{n=0}^{K_2-1} R_i^{(k)}(u-m, v-n) w_{j,i}^{(k)}(m,n) \right) + b_j^{(k)}, \quad (3)$$

$$k = 1, 2, ..., 21; j = 1, 2, ..., D_2; 0 \leq u \leq M_1 + K_1 - 1; 0 \leq v \leq M_2 + K_2 - 1,$$

where $D_1$ and $D_2$ are respectively the number of input and output feature maps, $K_1$ and $K_2$ are respectively the height and width of convolution kernel, and $b_j^{(k)}$ is the bias.

In most cases, we use small convolution kernels, such as 2×2, 3×3, and 4×4. For Branch #21, we use kernel size 1×1 in the first convolutional block to extract information among AC sub-bands. The specific kernel size settings depending on specific input can be referred to Table 1. Note that "SAME" padding is used in our JPEG-CNN, which means the output feature map has the same spatial dimension as the input feature map.

*Table1. The convolutional kernel size and output feature maps setting.*

| Branch | Conv1 Block | | Conv2 Block | | Conv3 Block | | Conv4 Block | |
|---|---|---|---|---|---|---|---|---|
| | Kernel size | Output Feature maps | Kernel size | Output Feature maps | Kernel size | Output Feature maps | Kernel size | Output Feature maps |
| Branch #1~ #20 | 3×3 | 2 | 2×2 | 4 | 4×4 | 4 | N/A | |
| Branch #21 | 1×1 | 2 | 3×3 | 4 | 2×2 | 4 | 4×4 | 4 |

- **Batch Normalization layer**

In order to better regularize the network, a batch normalization (BN) layer (Ioffe & Szegedy, 2015) follows the convolutional layer to normalize each feature map $F_{j,t}^{(k)}$ (referred to equation (3)) in *t-th* batch. Denote $E_{j,t}$ and $V_{j,t}$ the mean and the variance of $F_{j,t}^{(k)}$ for a mini-batch. Denote $\bar{E}_{j,t}$ and $\bar{V}_{j,t}$ the moving averaged mean and variance, which are updated in every batch through the follow equations

$$\bar{E}_{j,t} = \tau \times \bar{E}_{j,t-1} + (1-\tau) \times E_{j,t}, \quad j=1,2,...,D_2; \ t=2,3,...,T; \tag{4}$$

$$\bar{V}_{j,t} = \tau \times \bar{V}_{j,t-1} + (1-\tau) \times V_{j,t}, \quad j=1,2,...,D_2; \ t=2,3,...,T; \tag{5}$$

where $T$ is the total numbers of mini-batches, and $\tau$ is a momentum parameter. The normalization is performed as

$$B_{j,t}^{(k)} = \gamma_{j,t}^{(k)} \frac{F_{j,t}^{(k)} - \bar{E}_{j,t}}{\sqrt{\bar{V}_{j,t} + \xi}} + \beta_{j,t}^{(k)}, \quad k=1,2,...,21; j=1,2,...,D_2, \tag{6}$$

where $\gamma_{j,t}^{(k)}$ and $\beta_{j,t}^{(k)}$ are the scaling parameter and shifting parameter to be learned, respectively. The parameter $\xi$ is set as a small positive constant to prevent the denominator from being zero in this equation. By equation (6), each output feature map from previous convolutional layer is normalized in every batch.

- **TanH activation function**

Activation function is important to introduce non-linearity. It can increase the modeling capability of a network. There are a plenty of activation functions, including Sigmoid, TanH, ReLU, Leaky ReLU (Krizhevsky *et al.*, 2012), ELU (Clevert *et al.*, 2015), etc. We adopt TanH as the activation function in this network.

The TanH function is performed in an element-wise fashion by transforming the input $x$ to output $y$ with the following equation:

$$y = \frac{e^x - e^{-x}}{e^x + e^{-x}}. \tag{7}$$

- **Average pooling layer**

To maintain shift invariance while reducing the size of feature maps, a pooling layer is used. A sliding window size of 3×3 with stride 2 is used for each feature map.

- **Fully connected layer**

At the end of each branch, the feature maps are concatenated into a one dimensional vector, and then fed to two fully connected (FC) layers. Denote $a_m^l$ as the *m-th* neuron in the *l-th* layer. The output of the *n-th* neuron in the *(l+1)-th* layer is obtained as follows:

$$a_n^{l+1} = f\left(\sum_m a_m^l w_{m,n}^l + b_n^l\right), \tag{8}$$

where $w_{m,n}^l$ and $b_n^l$ are the weight and the bias to be learned, respectively, and $f(\cdot)$ is an activation function. In the first FC layer, TanH is used for activation. In the second FC layer, a Softmax function (normalized exponential function) is used to compute the *n-th* categorical probability as follows:

$$p_n = \frac{e^{a_n}}{\sum_{m=1}^C e^{a_m}}, \qquad n=1,2,...,C, \tag{9}$$

where *C* is the number of categories, and $a_m$ represents the output of the *m-th* neuron in the previous fully connected layer.

**Loss Function and Parameter Learning**

We use cross-entropy function (Murphy, 2012) as the loss function in the network. Given the actual output $p_n$ obtained by equation (9) and the expected ground truth output $q_n$, the loss is computed by:

$$L(w) = -\sum_{n=1}^C q_n \log(p_n). \tag{10}$$

Based on the loss function, the weights in the network architecture described above are learned by back propagation (BP) algorithm (Rumelhart *et al.*, 1988). Given the learning rate $\alpha$, the weight in the *t-th* iteration is updated towards the gradient direction as follows:

$$w^{(t)} = w^{(t-1)} - \alpha \cdot \nabla L(w), \tag{11}$$

where $\nabla L(w)$ is the derivative of the loss with respect to the weight.

## EXPERIMENTS

In this section, we report the performance of the proposed JPEG-CNN. The settings of the experiments are firstly introduced, including the software and hardware platform, the dataset, and the hyper-parameters. Then we present the experiments results on double JPEG compression. Finally, the effectiveness of the employed techniques, such as multiple branches, ABS layer, and BN layer, are investigated.

### Experimental Setups

We implement the proposed JPEG-CNN based on TensorFlow (Abadi *et al.*, 2016) version 1.1.0, an open-source platform for deep machine learning. Experiments are run on NVidiaTesla-P100 GPU with 16 GB memory.

A widely used image dataset, BOSSBase v1.01 (Bas *et al.*, 2011), is employed in the experiments. It contains 10,000 gray-scale images stored in uncompressed PGM (portable grey map) format with size 512×512. We augment the dataset as follows. First we randomly split the dataset with ratio 3:1 (7,500:2,500) for the training and testing phase. Then, each image is divided into four non-overlapping sub-images with size 256×256. This operation enlarges the number of images quadrupled (30,000:10,000). Finally, images are JPEG compressed once or twice in order to simulate the singly compressed images and doubly compressed images. The quality factor (QF) ranges from 60 to 95 with a step of 5 for both the primary compression (QF1) and the secondary compression (QF2).

The hyper-parameters and the training strategies used in JEPG-CNN are set as follows.

1) In convolutional layers, kernel parameters are initialized by random numbers following zero-mean Gaussian distribution with the standard deviation of 0.1. Bias parameters are initialed with uniform distribution ranged in [0, 1]. Dropout trick (Krizhevsky *et al.*, 2012) is not used.

2) In BN layers, the momentum in Equation (4) and (5) is set to 0.999, and the constant in Equation (6) is set to 0.01. The batch mean and variance of the best validation model are stored for testing.

3) In FC layers, weight parameters are initialized with the "Xavier" initialization (Glorot & Bengio, 2010), and bias parameters are initialized with zeros. Dropout trick is not used.

4) Training images are further divided into a training set and a validation set with the ratio 11:1. In other words, since the total number of training images are 60,000 (with 30,000 in respective class), we have 55,000 images for the training set and the rest 5,000 images for the validation set. All 55,000 images are run once in a training epoch. The network is trained by 80 epochs. A validation is performed after the 40-th epoch, and the best model in validation is saved for testing.

5) Stochastic gradient descent is used to train the JPEG-CNN with mini-batches. The batch size is set to 50, in which 25 are singly compressed images and the other 25 are the corresponding doubly compressed images. Therefore, an epoch contains 1100 iterations of batches. In each epoch, training images are shuffled to make different combinations of batches.

6) The learning rate is set to 0.05 initially, and it decreases 70% for every 20 epochs during training for all learned parameters.

The MBFDF (mode based first digit features) method together with the FLD (fisher linear discriminant) proposed in (Li *et al.*, 2008) is used for comparison. Although this handcrafted feature based method was proposed almost ten years ago, it is still an effective method used in many works (Amerini *et al.*, 2014; Milani *et al.*, 2014; Taimori *et al.*, 2016) and can achieve the state-of-the-art performance. Furthermore, the same as JPEG-CNN, the first 20 sub-bands AC coefficients in zigzag order are served as the input to obtain the FSD features.

**Learning Progress Visualization**

To visualize the learning progress of the proposed JPEG-CNN, Fig. 3 is depicted for demonstrating the accuracies for the training, validation, and testing sets, by taking the QF combination (QF1,QF2)=(60,65) for example. The classification accuracy is computed by the number of correctly classified singly compressed and double compressed images over the total number of images involved. As shown in the figure, the three curves share similar convergent trend, where the curve for validation and that for testing are closer. After the 40-th epoch, the accuracy of the validation set is rather stable. As a result, we use the best performed validated model after the 40-th epoch for testing.

*Figure 3. The accuracy of three sets for each learning epoch.*

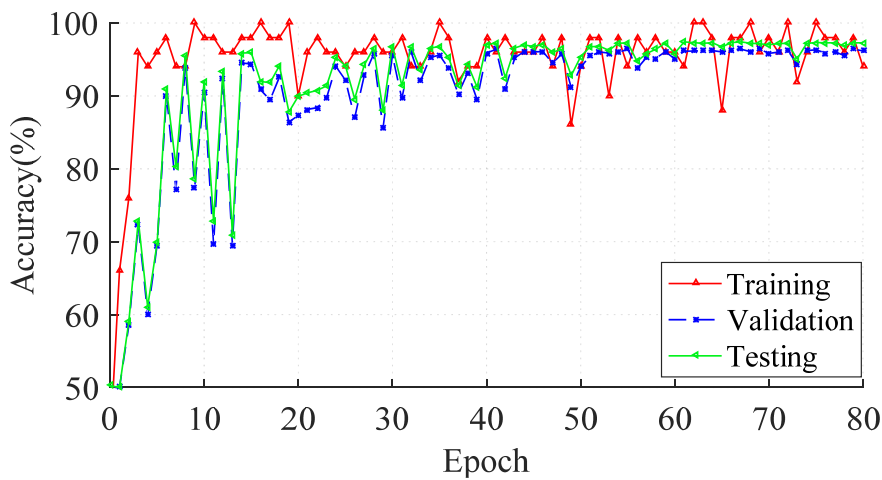

## Comparative Results

The performance of our proposed method and those of the comparative method are reported in Table 2 and 3, respectively. The performances are evaluated by the classification accuracy in the testing stage. We can observe that, in most cases, especially when QF2 is larger than QF1, the performance of JPEG-CNN is comparable with the handcrafted feature based method. Moreover, in those cases when QF1 is much greater than QF2, JPEG-CNN has advantage over the handcrafted features. For instance, in QF combination (QF1,QF2)=(95,65), the detecting accuracy has increased 4.61%. From the perspective of averaged accuracy with respective to QF2, JPEG-CNN shows slightly better performances than the MBFDF method.

## Effectiveness of Each Component

In order to better understand the proposed JPEG-CNN, we perform some experiments to investigate the impact of each network component. To this end, we remove or replace a specific component in the JPEG-CNN, and report the averaged accuracy for QF2, with QF1 ranging from 60 to 95 with a step of 5 (QF1 ≠ QF2).

Firstly, we investigate whether the multi-branch structure takes effect. In order to disable the multi-branch structure, we remove branches from Branch #1 to Branch #20, and leave Branch #21 only to take effect. Note that all DCT coefficients in the first 20 AC sub-bands are used as the input as the JPEG-CNN. The results are drawn in Fig. 4. We can observe that without the branches for processing intra sub-band information independently, the performance of only using the inter sub-band branch (Branch #21) drops to a large extent when QF2 is small. The results indicate that the proposed JPEG-CNN benefits from incorporating the domain knowledge by extracting features from individual sub-band.

Secondly, we remove all ABS layers in the network. The ABS layers in JPEG-CNN discard the sign information of the input DCT coefficients. The results are shown in Fig. 5. We can observe that without the ABS layers in the network, the redundant sign information may have a negative impact on the performance.

*Table 2.  The classification accuracy of the proposed JPEG-CNN.*

| QF2 / QF1 | 60 | 65 | 70 | 75 | 80 | 85 | 90 | 95 |
|---|---|---|---|---|---|---|---|---|
| 60 | —— | 98.53% | 99.70% | 99.57% | 99.83% | 99.89% | 99.96% | 99.99% |
| 65 | 97.88% | —— | 98.53% | 99.21% | 99.60% | 99.80% | 99.96% | 99.92% |
| 70 | 98.66% | 98.19% | —— | 98.44% | 99.61% | 99.86% | 99.97% | 99.98% |
| 75 | 99.05% | 98.13% | 96.96% | —— | 99.53% | 99.83% | 99.79% | 99.96% |
| 80 | 99.41% | 98.43% | 98.97% | 98.13% | —— | 99.64% | 99.91% | 99.98% |
| 85 | 96.49% | 98.91% | 99.17% | 98.73% | 99.10% | —— | 99.94% | 99.99% |
| 90 | 97.55% | 94.62% | 96.86% | 96.80% | 99.37% | 99.54% | —— | 99.85% |
| 95 | 80.43% | 87.92% | 84.08% | 92.17% | 94.02% | 97.49% | 99.65% | —— |
| Averaged | 95.64% | 96.39% | 96.32% | 97.58% | 98.72% | 99.44% | 99.88% | 99.95% |

*Table 3.  The classification accuracy of MBFDF method proposed in (Li et al., 2008).*

| QF2 / QF1 | 60 | 65 | 70 | 75 | 80 | 85 | 90 | 95 |
|---|---|---|---|---|---|---|---|---|
| 60 | —— | 99.46% | 99.73% | 99.96% | 99.99% | 99.98% | 99.98% | 99.95% |
| 65 | 98.71% | —— | 99.20% | 99.82% | 99.99% | 99.98% | 99.99% | 100.00% |
| 70 | 99.29% | 98.32% | —— | 99.22% | 99.96% | 99.97% | 99.98% | 100.00% |
| 75 | 99.16% | 99.09% | 98.77% | —— | 99.94% | 99.98% | 99.98% | 100.00% |
| 80 | 94.91% | 98.39% | 99.35% | 99.36% | —— | 99.92% | 99.98% | 100.00% |
| 85 | 98.40% | 97.97% | 97.70% | 99.42% | 99.59% | —— | 99.98% | 100.00% |
| 90 | 94.26% | 93.11% | 97.41% | 97.41% | 98.56% | 99.58% | —— | 99.98% |
| 95 | 76.20% | 83.32% | 79.48% | 86.78% | 90.88% | 97.66% | 99.24% | —— |
| Averaged | 94.42% | 95.66% | 95.95% | 97.42% | 98.41% | 99.58% | 99.87% | 99.99% |

Thirdly, we remove all BN layers in the network. The BN layers serve as regularizors to prevent internal covariate shift (Ioffe & Szegedy, 2015) and eliminate the need of using Dropout. The results are also shown in Fig. 5. We can observe that without the BN layers, the network distinctly performs worse.

Fourthly, we replace all TanH activiation functions with ReLU functions. Although ReLU performs faster and better than TanH in some computer vision tasks (Simonyan & Zisserman, 2014; He *et al.*, 2016), it has a truncation effect on the negative part of the input. As a result, it does not preserve all processed information. As shown in Fig. 6, the JPEG-CNN equipped with the TanH activation functions, which non-linearly map the input monotonously, achieves better performance than using ReLU.

Finally, we replace all average pooling operations with max pooling operations which are frequently used in many famous CNN structures, such as (Krizhevsky *et al.*, 2012; Szegedy *et al.*, 2015). The results are shown in Fig. 6, and indicate that although the average pooling operation may have higher computational complexity, it may preserve more discriminative information than max pooling in sub-sampling.

*Figure 4.    The performance of disabling multi-branch structure.*

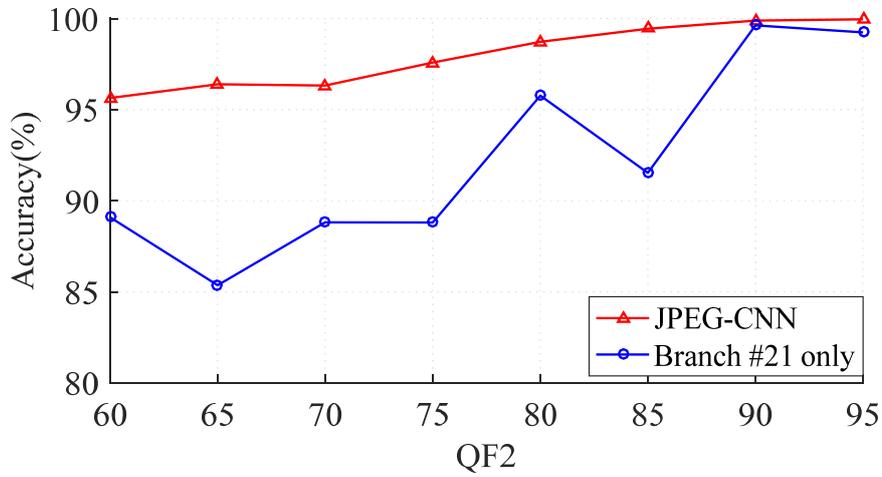

*Figure 5.    The performance of removing ABS layers or BN layers.*

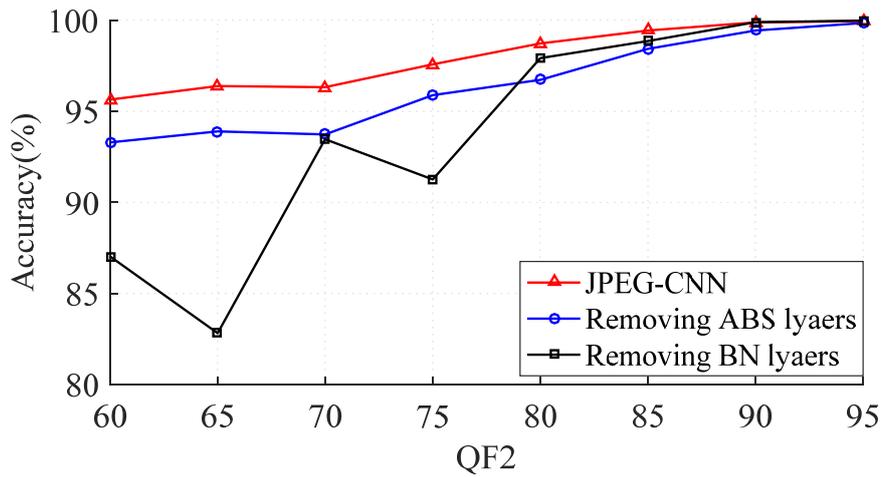

*Figure 6.    The performance of replacing TanH activation or average pooling operation.*

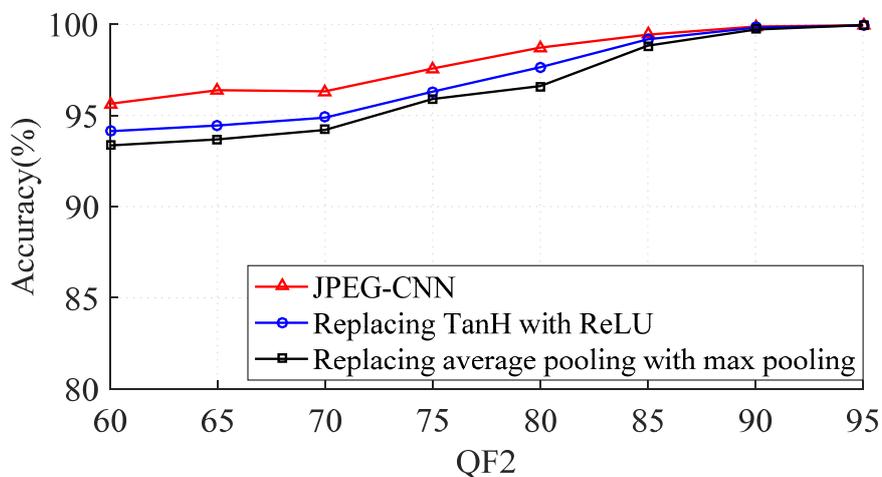

# CONCLUSION

In this paper, we propose a multi-branch CNN architecture, called JPEG-CNN. To the best of our knowledge, it may be the first CNN structure taking raw DCT coefficients as input, and adapting to JPEG DCT coefficients by considering their intricate relation within each sub-band and among all of the sub-bands. Through extensive experiments, we show that the proposed scheme provides end-to-end capability in detecting double JPEG compression. We also have experimentally verified that the ABS layer, BN layer, average pooling layer, and TanH function, all have contributed to the good performance of JPEG-CNN.

Although the proposed JPEG-CNN can effectively classify singly compressed images and doubly compressed images, the room for its improvement is still big. First, the network has only three or four convolutional blocks for each branch, and it may not be deep enough. It is well known that deep networks can represent certain function classes exponentially more efficiently than shallow ones. Therefore, a deeper network may be considered in the future. Some structures such as residual connection (He *et al.*, 2016) and dense connection (Huang *et al.*, 2016) may help. Second, the network takes constant sized image as input, which may require dividing an image of larger size into constant sized blocks in forensic applications. Therefore, a network that can take input with variable size should also be considered in future. Third, an experienced counterfeiter may develop anti-forensics/counter-forensics (Stamm & Liu, 2011) operations to avoid DQ effect. It is appealing to study the network performance on detecting counter-forensics operations (Lai & Böhme, 2011; Barni *et al.*, 2017) and make corresponding improvement in the network structure.

Xie, S., Girshick, R., Dollár, P., Tu, Z., & He, K. (2016). Aggregated residual transformations for deep neural networks. *arXiv preprint arXiv:1611.05431*.